\documentclass{article}

\usepackage{microtype}
\usepackage{graphicx}
\usepackage{subfigure}
\usepackage{multirow}
\usepackage{pifont}
\usepackage{booktabs} 

\usepackage{amsmath,amsfonts,bm}

\def\eqref#1{(\ref{#1})}

\def\1{\bm{1}}

\DeclareMathAlphabet{\mathsfit}{\encodingdefault}{\sfdefault}{m}{sl}
\SetMathAlphabet{\mathsfit}{bold}{\encodingdefault}{\sfdefault}{bx}{n}

\newcommand{\Def}[0]{\mathrel{\mathop:}=}
\usepackage[dvipsnames]{xcolor}

\usepackage{hyperref}

\usepackage[accepted]{icml2021}

\usepackage{amsmath}
\usepackage{amsthm}
\usepackage{amssymb}
\usepackage{enumitem}
\definecolor{Tianlong_color}{rgb}{0.858, 0.188, 0.478}

\renewcommand{\algorithmicrequire}{ \textbf{Input:}} 
\renewcommand{\algorithmicensure}{ \textbf{Output:}} 

\icmltitlerunning{A Unified Lottery Ticket Hypothesis for Graph Neural Networks}

\begin{document}

\twocolumn[
\icmltitle{A Unified Lottery Ticket Hypothesis for Graph Neural Networks}



\icmlsetsymbol{equal}{*}

\begin{icmlauthorlist}
\icmlauthor{Tianlong Chen}{equal,ut}
\icmlauthor{Yongduo Sui}{equal,ustc}
\icmlauthor{Xuxi Chen}{ut}
\icmlauthor{Aston Zhang}{ama}
\icmlauthor{Zhangyang Wang}{ut}
\end{icmlauthorlist}

\icmlaffiliation{ut}{Department of Electrical and Computer Engineering, University of Texas at Austin}
\icmlaffiliation{ustc}{University of Science and Technology of China}
\icmlaffiliation{ama}{AWS Deep Learning}
\icmlcorrespondingauthor{Zhangyang Wang}{atlaswang@utexas.edu}

\icmlkeywords{Machine Learning, ICML}

\vskip 0.3in
]



\printAffiliationsAndNotice{\icmlEqualContribution} 

\begin{abstract}
With graphs rapidly growing in size and deeper graph neural networks (GNNs) emerging, the training and inference of GNNs become increasingly expensive. Existing network weight pruning algorithms cannot address the main space and computational bottleneck in GNNs, caused by the size and connectivity of the graph. To this end, this paper first presents a unified GNN sparsification (UGS) framework that simultaneously prunes the graph adjacency matrix and the model weights, for effectively accelerating GNN inference on large-scale graphs. Leveraging this new tool, we further generalize the recently popular \emph{lottery ticket hypothesis} to  GNNs for the first time, by defining a \emph{graph lottery ticket} (GLT) as a pair of core sub-dataset and sparse sub-network, which can be jointly identified from the original GNN and the full dense graph by iteratively applying UGS. Like its counterpart in convolutional neural networks, GLT can be trained in isolation to match the performance of training with the full model and graph, and can be drawn from both randomly initialized and self-supervised pre-trained GNNs. Our proposal has been experimentally verified across various GNN architectures and diverse tasks, on both small-scale graph datasets (Cora, Citeseer and PubMed), and large-scale datasets from the challenging Open Graph Benchmark (OGB). Specifically, for node classification, our found GLTs achieve the same accuracies with $20\%\sim98\%$ MACs saving on small graphs and $25\%\sim85\%$ MACs saving on large ones. For link prediction, GLTs lead to $48\%\sim97\%$ and $70\%$ MACs saving on small and large graph datasets, respectively, without compromising predictive performance. Codes are at {\small\url{https://github.com/VITA-Group/Unified-LTH-GNN}}.
\end{abstract}

\section{Introduction}
Graph Neural Networks (GNNs) \cite{zhou2018graph,kipf2016semi,chen2019equivalence,velivckovic2017graph} have established state-of-the-art results on various graph-based learning tasks, such as node or link classification \cite{kipf2016semi,velivckovic2017graph,qu2019gmnn,verma2019graphmix,karimi2019explainable,you2020l2,pmlr-v119-you20a}, link prediction \cite{zhang2018link}, and graph classification \cite{ying2018hierarchical,xu2018powerful,you2020graph}. GNNs' superior performance results from the structure-aware exploitation of graphs. To update the feature of each node, GNNs first aggregate features from neighbor connected nodes, and then transform the aggregated embeddings via (hierarchical) feed-forward propagation. 

\begin{figure}[t]
 \vspace{-1em}
    \centering
    \includegraphics[width=1\linewidth]{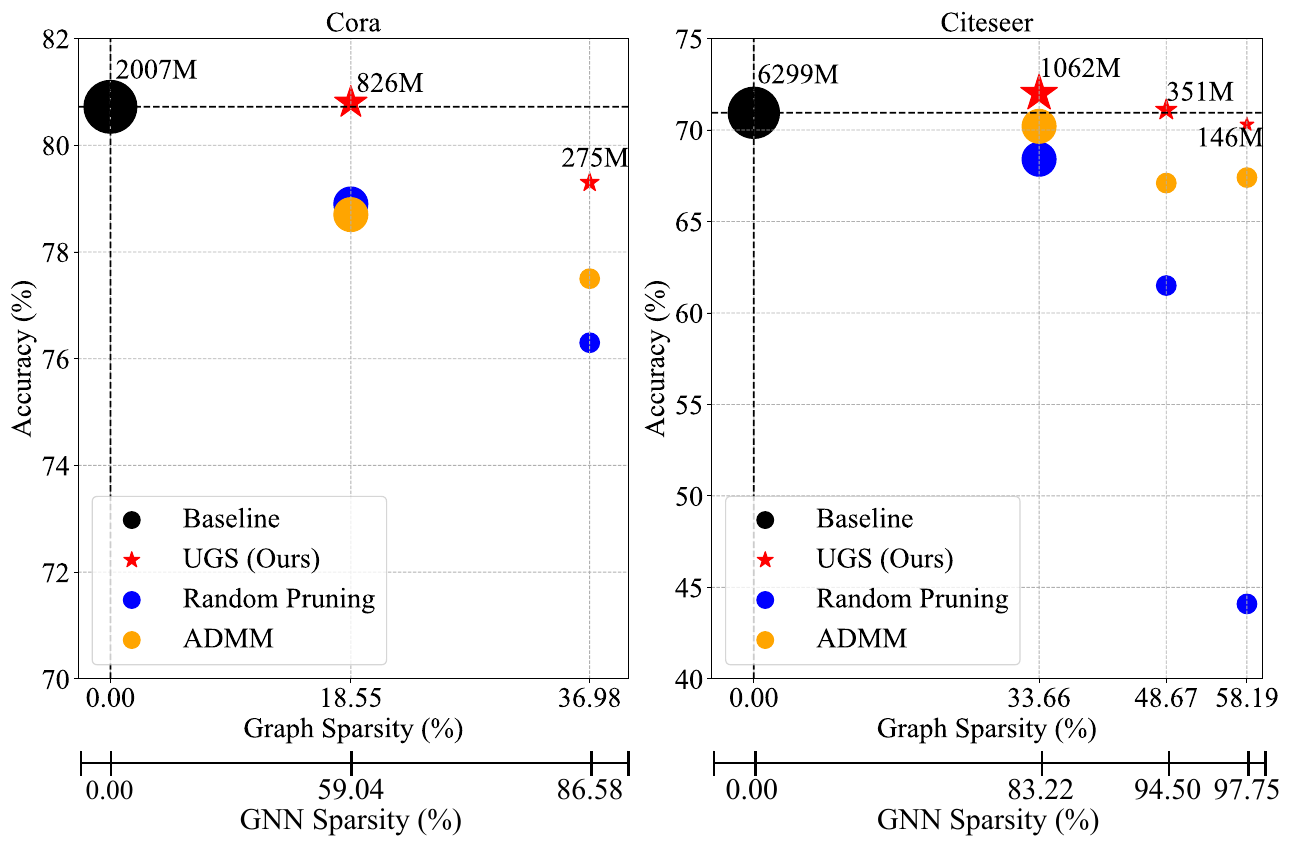}
    \vspace{-1.5em}
    \caption{Summary of our achieved performance (y-axis) at different graph and GNN sparsity levels (x-axis) on Cora and Citeceer node classification. The \textit{size of markers} represent the inference MACs ($=\frac{1}{2}$ FLOPs) of each sparse GCN on the corresponding sparsified graphs. \textit{Black circles} ($\bullet$) indicate the baseline, i.e., unpruned dense GNNs on the full graph. \textit{Blue circles} ($\textcolor{blue}{\bullet}$) are random pruning results. \textit{Orange circles} ($\textcolor{orange}{\bullet}$) represent  the performance of a previous graph sparsification approach, i.e., ADMM \cite{li2020sgcn}. \textit{Red stars} (\textcolor{red}{\ding{72}}) are established by our method (UGS).}  
     \vspace{-1em}
    \label{fig:teaser}
\end{figure}

However, the training and inference of GNNs suffer from the notorious inefficiency, and pose hurdle to GNNs from being scaled up to real-world large-scale graph applications. This hurdle arises from both algorithm and hardware levels. \underline{On the algorithm level}, GNN models can be thought of a composition of traditional graphs equipped with deep neural network (DNN) algorithms on vertex features. The execution of GNN inference falls into three distinct categories with unique computational characteristics: graph traversal, DNN computation, and aggregation. Especially, GNNs broadly follow a recursive neighborhood aggregation (or message passing) scheme, where each node aggregates feature vectors of its multi-hop neighbors to compute its new feature vector. The aggregation phase costs massive computation when the graphs are large and with dense/complicated neighbor connections \cite{xu2018powerful}. \underline{On the hardware level}, GNN's computational structure depends on the often sparse and irregular structure of the graph adjacency matrices. This results in many random memory accesses and limited data reuse, but also requires relatively little computation.  As a result, GNNs have much higher inference latency than other neural networks, limiting them to applications where inference can be pre-computed offline  \cite{geng2020awb,yan2020hygcn}.



This paper aims at aggressively trimming down the explosive GNN complexity, from the algorithm level. There are two streams of works: \textit{simplifying the graph}, or \textit{simplifying the model}. For \underline{the first} stream, many have explored various sampling-based strategies  \cite{hubler2008metropolis,chakeri2016spectral,calandriello2018improved,adhikari2017propagation,leskovec2006sampling,voudigari2016rank,eden2018provable,zhao2015gsparsify,chen2018fastgcn}, often combined with mini-batch training algorithms for locally aggregating and updating features. \citet{zheng2020robust} investigated graph sparsification, i.e., pruning input graph edges, and learned an extra DNN surrogate. \citet{li2020sgcn} also addressed graph sparsification by formulating an optimization objective, solved by alternating direction method of multipliers (ADMM) \cite{bertsekas1982constrained}. 

\underline{The second} stream of efforts were traditionally scarce, since the DNN parts of most GNNs are (comparably) lightly parameterized, despite the recent emergence of increasingly deep GNNs \cite{li2019deepgcns}. Although model compression is well studied for other types of DNNs \cite{cheng2017survey}, it has not been discussed much for GNNs. One latest work \cite{tailor2021degreequant} explored the viability of training quantized GNNs, enabling the usage of low precision integer arithmetic during inference. Other forms of well-versed DNN compression techniques, such as model pruning \cite{han2015deep}, have not been exploited for GNNs up to our best knowledge. More importantly, no prior discussion was placed on jointly simplifying the input graphs and the models for GNN inference. In view of such, this paper asks: \textit{to what extent could we co-simplify the input graph and the model, for ultra-efficient GNN inference}?

\vspace{-0.5em}
\subsection{Summary of Our Contributions}
\vspace{-0.2em}
This paper makes multi-fold contributions to answer the above questions. Unlike pruning convolutional DNNs which are heavily overparameterized, directly pruning the much less parameterized GNN model would have only limited room to gain. \textbf{Our first technical innovation} is to for the first time present an end-to-end optimization framework called \textit{unified GNN sparsification} (\textbf{UGS}) that simultaneously prunes the graph adjacency matrix and the model weights. UGS makes no assumption to any GNN architecture or graph structure, and can be flexibly applied across various graph-based learning scenarios at scale.


Considering UGS as the generalized pruning for GNNs, \textbf{our second technical innovation} is to generalize the popular \textit{lottery ticket hypothesis} (\textbf{LTH}) to GNNs for the first time. LTH \cite{frankle2018lottery} demonstrates that one can identify highly sparse and independently trainable subnetworks from dense models, by iterative pruning. It was initially observed in convolutional DNNs, and later broadly found in natural language processing (NLP) \cite{chen2020lottery}, generative models \cite{kalibhat2020winning}, reinforcement learning \cite{yu2019playing} and lifelong learning \cite{chen2020lottery}. To meaningfully generalize LTH to GNNs, we define a \textit{graph lottery ticket} (\textbf{GLT}) as \textit{a pair of core sub-dataset and sparse sub-network} which can be jointly identified from  the full graph and the original GNN model, by iteratively applying UGS. Like its counterpart in convolutional DNNs, a GLT could be trained from its initialization to match the performance of training with the full model and graph, and its inference cost is drastically smaller.

Our proposal has been experimentally verified, across various GNN architectures and diverse tasks, on both small-scale graph datasets (Cora, Citeseer and PubMed), and large-scale datasets from the challenging Open Graph Benchmark (OGB). \textbf{Our main observations} are outlined below:
\begin{itemize}\vspace{-1em}
    \item UGS is widely applicable to simplifying a GNN during training and reducing its inference MACs (multiply–accumulate operations). Moreover, by iteratively applying UGS, GLTs can be broadly located from for both shallow and deep GNN models, on both small- and large-scale graph datasets, with substantially reduced inference costs and unimpaired generalization.\vspace{-0.3em}
    \item For node classification, our found GLTs achieve $20\%\sim98\%$ MACs saving, with up to $5\%\sim58.19\%$ sparsity on graphs and $20\%\sim97.75\%$ sparsity on GNN models, at little to no performance degradation. For example in Figure~\ref{fig:teaser}, on Cora and Citeseer node classification, our GLTs (\textcolor{red}{\ding{72}}) achieve comparable or sometimes even slightly better performance than the baselines of full models and graphs ($\textcolor{black}{\bullet}$),  with only $41.16\%$ and $5.57\%$ MACs, respectively.\vspace{-0.3em} 
    \item  For link prediction, GLTs lead to $48\%\sim97\%$ and $70\%$ MACs saving, coming from up to $22.62\%\sim55.99\%$ sparsity on graphs and and $67.23\%\sim97.19\%$ sparsity on GNN models, again without performance loss.\vspace{-0.3em}
    \item Our proposed framework can scale up to deep GNN models (up to $28$ layers) on large graphs (e.g., Ogbn-ArXiv and Ogbn-Proteins), without bells and whistles.\vspace{-0.3em}
    \item Besides from random initializations, GLTs can also be drawn from the initialization via self-supervised pre-training -- an intriguing phenomenon recently just reported for NLP \cite{chen2020lottery} and computer vision models \cite{chen2020lottery2}. Using a latest GNN pre-training algorithm \cite{you2020graph} for initialization, GLTs can be found to achieve robust performance with even sparser graphs and GNNs.\vspace{-0.5em}
\end{itemize}
\vspace{-0.2em}
\section{Related Work}
\vspace{-0.2em}
\paragraph{Graph Neural Networks.} 
There are mainly three categories of GNNs \cite{dwivedi2020benchmarking}: i) extending original convolutional neural networks to the graph regime~\citep{scarselli2008graph,bruna2013spectral,kipf2016semi,hamilton2017inductive}; ii) introducing anisotropic operations on graphs such as gating and attention~\citep{battaglia2016interaction,monti2017geometric,velivckovic2018graph}, and iii) improving upon limitations of existing models~\citep{xu2018how,morris2019weisfeiler,chen2019equivalence,murphy2019relational}. Among this huge family, Graph Convolutional Networks (GCNs) are widely adopted, which can be categorized as spectral domain based methods \cite{defferrard2016convolutional, kipf2016semi} and spatial domain bases methods \cite{simonovsky2017dynamic, hamilton2017inductive}. 


The computational cost and memory usage of GNNs will expeditiously increase with the graph size. The aim of graph sampling or sparsification is to extract a small sub-graph from the original large one, which can remain effective for learning tasks~\citep{zheng2020robust,hamilton2017inductive} while reducing the cost. Previous works on sampling focus on preserving certain pre-defined graph metrics~\citep{hubler2008metropolis}, graph spectrum~\citep{chakeri2016spectral,adhikari2017propagation}, or node distribution~\citep{leskovec2006sampling,voudigari2016rank,eden2018provable}.
FastGCN~\citep{chen2018fastgcn} introduced a global importance sampling method instead of locally neighbor sampling.
VRGCN~\citep{chen2018stochastic} proposed a control variate based algorithm, but requires all intermediate vertex embeddings to be saved during training. Cluster-GCN~\citep{chiang2019cluster} used clustering to partition subgraphs for training, but often suffers in stability. \citet{zheng2020robust,li2020sgcn} cast graph sparsification as optimization problems, solved by learning surrogates and ADMM, respectively. 

\vspace{-1em}
\paragraph{Lottery Ticket Hypothesis (LTH).} Since the original LTH \citep{frankle2018lottery}, a lot of works have explored the prospect of
trainable sparse subnetworks in place of the full models without sacrificing performance. \citet{frankle2019linear,renda2020comparing} introduced the rewinding techniques to scale up LTH. LTH was also adopted in different fields \citep{evci2019difficulty, savarese2020winning,liu2018rethinking, You:2019tz,gale2019state,yu2019playing,kalibhat2020winning,chen2021long,chen2020lottery,chen2020earlybert,chen2021ultra,ma2021good,gan2021playing}. 

However, GNN is NOT ``yet another" field that can be easily cracked by LTH. That is again due to GNNs having much smaller models, while all the aforementioned LTH works focus on simplifying their redundant models. To our best knowledge, this work is not only the first to generalize LTH to GNNs, but also the first to extend LTH from simplifying models to a new data-model co-simplification prospect.



\begin{figure*}[t] 
\centering
\includegraphics[width=0.87\linewidth]{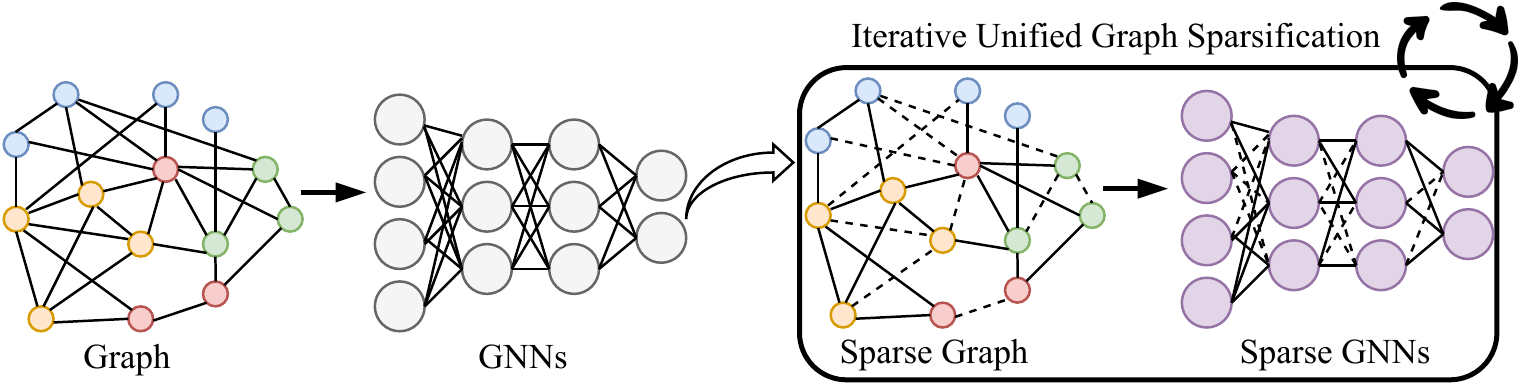}
\vspace{-3mm}
\caption{An illustration of unified GNN sparsification (UGS). Dash/solid lines denote the removed/remaining edges and weights in the graph and GNNs, respectively. Note that graphs and GNNs are co-optimized to find optimal solutions for the unified sparsification.}
\label{fig:UGS}
\vspace{-4mm}
\end{figure*}

\vspace{-0.5em}
\section{Methodology}
\vspace{-0.2em}
\subsection{Notations and Formulations}
\vspace{-0.3em}
Let $\mathcal{G} = \{\mathcal{V}, \mathcal{E}\}$ represent an undirected graph with $|\mathcal{V}|$ nodes and $|\mathcal{E}|$ edges. For $\mathcal{V} = \{v_1, ..., v_{|\mathcal{V}|}\}$, let $\boldsymbol{X} \in \mathbb{R}^{|\mathcal{V}| \times \mathrm{F}}$ denote the node feature matrix of the whole graph, where $\boldsymbol{x}_i = \boldsymbol{X}[i,:]$ is the $\mathrm{F}$-dimensional attribute vector of node $v_i\in\mathcal{V}$. As for $\mathcal{E} = \{e_1, ..., e_{\mathcal{|E|}}\}$, $e_n=(v_i, v_j)\in\mathcal{E}$ means that there exists a connection between node $v_i$ and $v_j$. An adjacency matrix $\boldsymbol{A} \in \mathbb{R}^{|\mathcal{V}| \times |\mathcal{V}| }$ is defined to describe the overall graph topology, where $\boldsymbol{A}[i,j]=1$ if $(v_i, v_j) \in \mathcal{E}$ else $\boldsymbol{A}[i,j]=0$. For example, the two-layer GNN \cite{kipf2016semi} can be defined as follows:
\begin{equation}
    \boldsymbol{Z} = \mathcal{S}(\boldsymbol{\hat{A}} \sigma(\boldsymbol{\hat{A}}\boldsymbol{X}\boldsymbol{\Theta}^{(0)})\boldsymbol{\Theta}^{(1)}),
\end{equation}
where $\boldsymbol{Z}$ is the prediction of GNN $f(\mathcal{G},\boldsymbol{\Theta})$. The graph $\mathcal{G}$ can be alternatively denoted as $\{\boldsymbol{A}, \boldsymbol{X}\}$, $\boldsymbol{\Theta}=(\boldsymbol{\Theta}^{(0)},\boldsymbol{\Theta}^{(1)})$ is the weights of the two-layer GNN, $\mathcal{S}(\cdot)$ represents the softmax function, $\sigma(\cdot)$ denotes the activation function (e.g., ReLU), $\boldsymbol{\hat{A}}=\boldsymbol{\tilde{D}}^{-\frac{1}{2}}(\boldsymbol{A}+\boldsymbol{I})\boldsymbol{\tilde{D}}^{\frac{1}{2}}$ is normalized by the degree matrix $\boldsymbol{\tilde{D}}$ of $\boldsymbol{A}+\boldsymbol{I}$. Considering the transductive semi-supervised classification task, the objective function $\mathcal{L}$ is:
\begin{equation}
    \mathcal{L}(\mathcal{G},\boldsymbol{\Theta}) = -\frac{1}{|\mathcal{V}_{\mathrm{label}}|} \sum_{v_i\in\mathcal{V}_{\mathrm{label}}}\boldsymbol{y}_i\mathrm{log}(\boldsymbol{z}_i),
\end{equation}
where $\mathcal{L}$ is the cross-entropy loss over all labeled samples $\mathcal{V}_{\mathrm{label}}\subset\mathcal{V}$, and $\boldsymbol{y}_i$ is the annotated label vector of node $v_i$ for its corresponding prediction $\boldsymbol{z}_i=\boldsymbol{Z}[i,:]$.

\vspace{-0.3em}
\subsection{Unified GNN Sparsification} \label{sec:ugs}
\vspace{-0.3em}
We present out end-to-end framework, \textbf{U}nified \textbf{G}NN \textbf{S}parsification (UGS), to simultaneously reduce edges in $\mathcal{G}$ and the parameters in GNNs. Specifically, we introduce two differentiable masks $\boldsymbol{m}_g$ and $\boldsymbol{m}_{\theta}$ for indicating the insignificant connections and weights in the graph and GNNs, respectively. The shapes of $\boldsymbol{m}_g$ and $\boldsymbol{m}_{\theta}$ are identical to those the adjacency matrix $\boldsymbol{A}$ and the weights $\boldsymbol{\Theta}$, respectively. Given $\boldsymbol{A}$, $\boldsymbol{\Theta}$, $\boldsymbol{m}_g$ and $\boldsymbol{m}_{\theta}$ are co-optimized from end to end, under the following objective:
\begin{align}
    \begin{array}{ll}
    \mathcal{L}_{\mathrm{UGS}} \Def & \mathcal{L}(\{\boldsymbol{m}_g\odot\boldsymbol{A},\boldsymbol{X}\},\boldsymbol{m}_{\theta}\odot\boldsymbol{\Theta}) \\ & + \gamma_1\|\boldsymbol{m}_g\|_{1} + \gamma_2\|\boldsymbol{m}_{\theta}\|_{1}, \label{eq:ugs}
\end{array}
\end{align}
where $\odot$ is the element-wise product, $\gamma_1$ and $\gamma_2$ are the hyparameters to control $\ell_1$ sparsity regularizers of $\boldsymbol{m}_g$ and $\boldsymbol{m}_{\theta}$ respectively. 
After the training is done, we set the lowest-magnitude elements in $\boldsymbol{m}_g$ and $\boldsymbol{m}_{\theta}$ to zero, w.r.t. pre-defined ratios $p_g$ and $p_{\theta}$. Then, the two sparse masks are applied to prune $\boldsymbol{A}$ and $\boldsymbol{\Theta}$, leading to the final sparse graph and model.
Alg.~\ref{alg1} outlines the procedure of UGS, and it can be considered as the generalized pruning for GNNs. 


\begin{algorithm}[t]
\caption{Unified GNN Sparsification (UGS)} 
\label{alg1}
\begin{algorithmic}[1]
\REQUIRE Graph $\mathcal{G} = \{\boldsymbol{A},\boldsymbol{X}\}$, GNN $f(\mathcal{G},\boldsymbol{\Theta}_0)$, GNN's initialization $\boldsymbol{\Theta}_0$, initial masks $\boldsymbol{m}_g^0 = \boldsymbol{A}$, $\boldsymbol{m}_{\theta}^0=\boldsymbol{1}\in\mathbb{R}^{\|\Theta_0\|_0}$, Step size $\eta$, $\lambda_g$, and $\lambda_{\theta}$.
\ENSURE Sparsified masks $\boldsymbol{m}_g$ and $\boldsymbol{m}_{\theta}$ 
\FOR{iteration $i=0,1,2,...,\mathrm{N}-1$}
\STATE  Forward $f(\cdot, \boldsymbol{m}_{\theta}^i \odot \boldsymbol{\Theta}_i)$ with $\mathcal{G}=\{\boldsymbol{m}_g^i \odot \boldsymbol{A}, \boldsymbol{X}\}$ to compute the loss $\mathcal{L}_\mathrm{UGS}$ in Equation~\ref{eq:ugs}.
\STATE  Backpropagate to update $\boldsymbol{\Theta}_{i + 1} \leftarrow \boldsymbol{\Theta}_i-\eta \nabla_{\boldsymbol{\Theta}_i}\mathcal{L}_\mathrm{UGS}$.
\STATE  Update $\boldsymbol{m}_g^{i+1} \leftarrow \boldsymbol{m}_g^{i}-\lambda_g\nabla_{\boldsymbol{m}_g^i}\mathcal{L}_\mathrm{UGS}$.
\STATE  Update $\boldsymbol{m}_{\theta}^{i+1} \leftarrow \boldsymbol{m}_{\theta}^{i}-\lambda_{\theta}\nabla_{\boldsymbol{m}_{\theta}^i}\mathcal{L}_\mathrm{UGS}$.
\ENDFOR
\STATE Set $p_g=5\%$ of the lowest magnitude values in $\boldsymbol{m}_g^{\mathrm{N}}$ to $0$ and others to $1$, then obtain $\boldsymbol{m}_g$.
\STATE Set $p_{\theta}=20\%$ of the lowest magnitude values in $\boldsymbol{m}_{\theta}^{\mathrm{N}}$ to $0$ and others to $1$, then obtain $\boldsymbol{m}_{\theta}$.
\end{algorithmic}
\end{algorithm}

\begin{algorithm}[t]
\caption{Iterative UGS to find Graph Lottery Tickets} 
\label{alg2}
\begin{algorithmic}[1]
\REQUIRE Graph $\mathcal{G} = \{\boldsymbol{A}, \boldsymbol{X}\}$, GNN $f(\mathcal{G},\boldsymbol{\Theta}_0)$, GNN's initialization $\boldsymbol{\Theta}_0$, pre-defined sparsity levels $s_g$ for graphs and $s_{\theta}$ for GNNs, Initial masks $\boldsymbol{m}_g = \boldsymbol{A}$, $\boldsymbol{m}_{\theta}=\boldsymbol{1}\in\mathbb{R}^{\|\Theta_0\|_0}$.
\ENSURE GLT $f(\{\boldsymbol{m}_g\odot\boldsymbol{A},\boldsymbol{X}\},\boldsymbol{m}_{\theta}\odot\boldsymbol{\Theta}_0)$
\WHILE {$1-\frac{\|\boldsymbol{m}_g\|_0}{\|\boldsymbol{A}\|_0} < s_g$ \textbf{and} $1-\frac{\|\boldsymbol{m}_{\theta}\|_0}{\|\boldsymbol{\Theta}\|_0} < s_{\theta}$}
\STATE Sparsify GNN $f(\cdot,\boldsymbol{m}_{\theta} \odot \boldsymbol{\Theta}_0)$ with $\mathcal{G}=\{\boldsymbol{m}_g \odot \boldsymbol{A}, X\}$ using UGS, as presented in Algorithm~\ref{alg1}.
\STATE Update $\boldsymbol{m}_g$ and $\boldsymbol{m}_{\theta}$ accordingly.
\STATE Rewinding GNN's weights to $\boldsymbol{\Theta}_0$
\STATE Rewinding masks, $\boldsymbol{m}_g=\boldsymbol{m}_g\odot\boldsymbol{A}$
\ENDWHILE 
\end{algorithmic}
\end{algorithm}

\vspace{-0.3em}
\subsection{Graph Lottery Tickets}

\paragraph{Graph lottery tickets (GLT).} Given a GNN $f(\cdot,\boldsymbol{\Theta})$ and a graph $\mathcal{G}=\{\boldsymbol{A},\boldsymbol{X}\}$, the associated subnetworks of GNN and sub-graph can be defined as $f(\cdot,\boldsymbol{m}_{\theta}\odot\boldsymbol{\Theta})$ and $\mathcal{G}_s=\{\boldsymbol{m}_g\odot\boldsymbol{A},\boldsymbol{X}\}$, where $\boldsymbol{m}_g$ and $\boldsymbol{m}_{\theta}$ are binary masks defined in Section~\ref{sec:ugs}. If a subnetwork $f(\cdot,\boldsymbol{m}_{\theta}\odot\boldsymbol{\Theta})$ trained on a sparse graph $\mathcal{G}_s$ has performance matching or surpassing the original GNN trained on the full graph $\mathcal{G}$ in terms of achieved standard testing accuracy, then we define $f(\{\boldsymbol{m}_g\odot\boldsymbol{A},\boldsymbol{X}\},\boldsymbol{m}_{\theta}\odot\boldsymbol{\Theta}_0)$ as a unified \textit{graph lottery tickets} (GLTs), where $\boldsymbol{\Theta}_0$ is the original initialization for GNNs which the found lottery ticket subnetwork is usually trained from.

Unlike previous LTH literature \cite{frankle2018lottery}, our identified GLT will consist of three elements: i) a sparse graph $\mathcal{G}_s=\{\boldsymbol{m}_g\odot\boldsymbol{A},\boldsymbol{X}\}$; ii) the sparse mask $\boldsymbol{m}_{\theta}$ for the model weight; and iii) the model weight's initialization $\boldsymbol{\Theta_0}$.

\vspace{-0.5em}
\paragraph{Finding GLT.} Classical LTH leverages iterative magnitude-based pruning (IMP) to identify lottery tickets. In a similar fashion, we apply our UGS algorithm to prune both the model and the graph during training, as outlined Algorithm~\ref{alg2}, obtaining the graph mask $\boldsymbol{m}_g$ and model weight mask $\boldsymbol{m}_{\theta}$ of GLT. Then, the GNN weights are rewound to the original initialization $\boldsymbol{\Theta}$. We repeat the above two steps iteratively, until reaching the desired sparsity $s_g$ and $s_{\theta}$ for the graph and GNN, respectively.

\vspace{-0.5em}
\paragraph{Complexity analysis of GLTs.} The inference time complexity of GLTs is $\mathcal{O}(\mathrm{L}\times\|\boldsymbol{m}_g\odot\boldsymbol{A}\|_0\times\mathrm{F}+\mathrm{L}\times\|m_{\theta}\|_0\times|\mathcal{V}|\times\mathrm{F}^2)$, where $\mathrm{L}$ is the number of layers, $\|\boldsymbol{m}_g\odot\boldsymbol{A}\|_0$ is the number of remaining edges in the sparse graph, $\mathrm{F}$ is the dimension of node features, $|\mathcal{V}|$ is the number of nodes. The memory complexity is $\mathcal{O}(\mathrm{L}\times|\mathcal{V}|\times\mathrm{F}+\mathrm{L}\times\|m_{\theta}\|_0\times\mathrm{F}^2)$. In our implementation, pruned edges will be removed from $\mathcal{E}$, and would not participate in the next round's computation. 

\begin{figure*}[!ht] 
\centering
\includegraphics[width=0.98\linewidth]{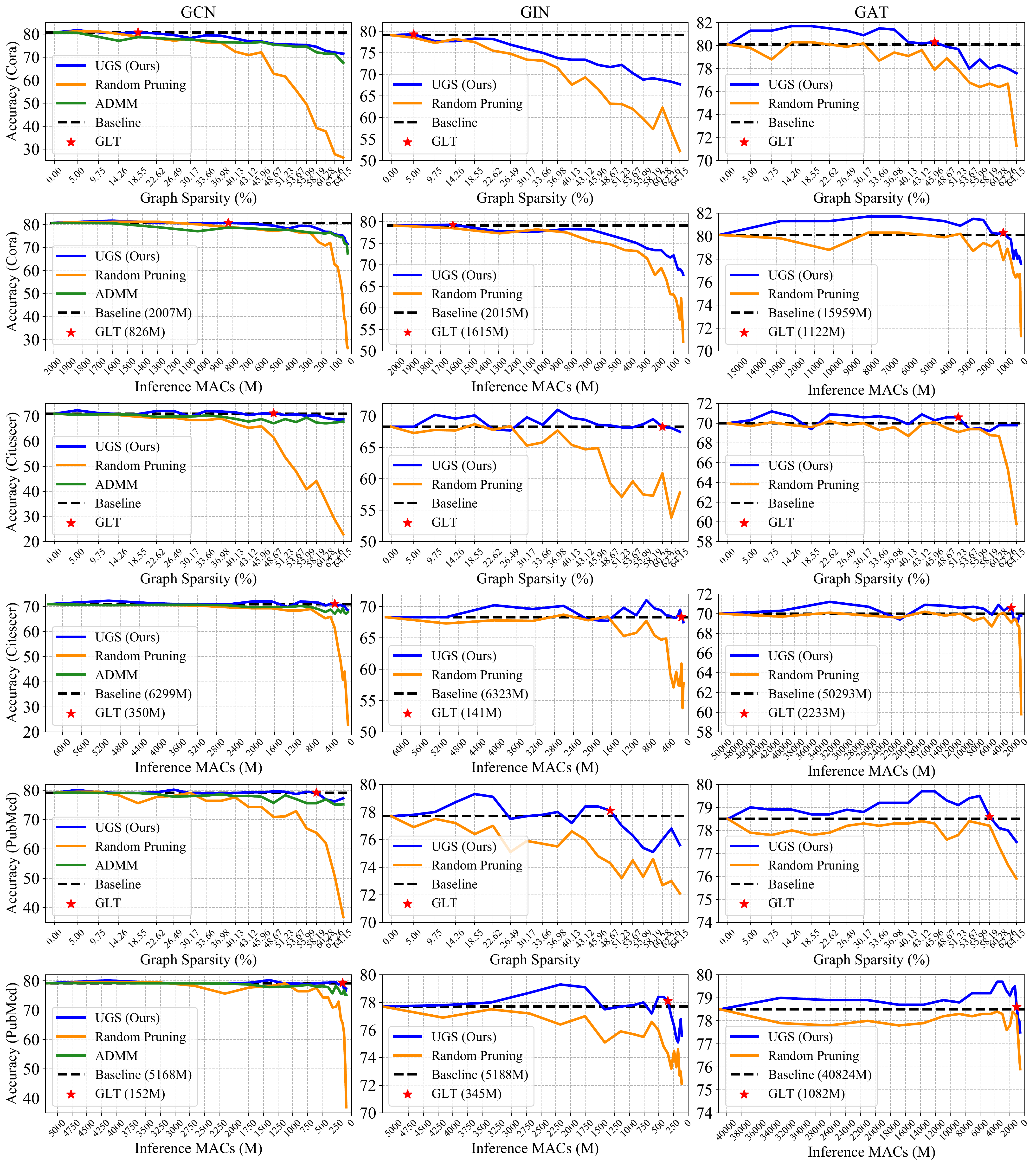}
\vspace{-5mm}
\caption{\textbf{Node classification} performance over achieved graph sparsity levels or inference MACs of GCN, GIN, and GAT on Cora, Citeseer, and PubMed datasets, respectively. \textit{Red stars} (\textcolor{red}{\ding{72}}) indicate the located GLTs, which reach comparable performance with high sparsity and low inference MACs. \textit{Dash lines} represent the baseline performance of full GNNs on full graphs. More results over GNN sparsity are provided in Appendix~\ref{sec:more_small_node}.}
\label{fig:node_cls}
\end{figure*}

\begin{figure*}[!ht] 
\centering
\includegraphics[width=0.98\linewidth]{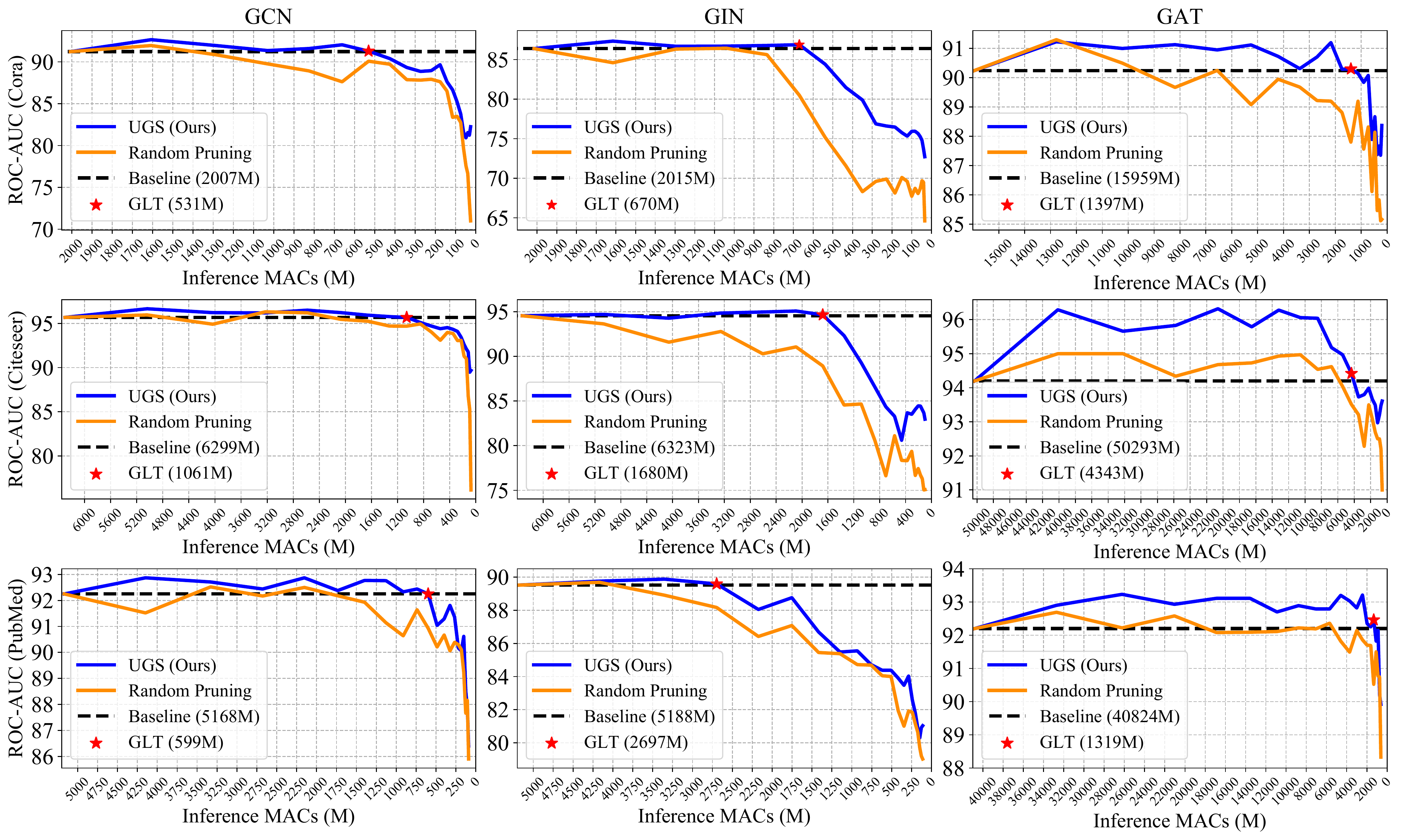}
\vspace{-5mm}
\caption{\textbf{Link prediction} performance over inference MACs of GCN, GIN, and GAT on Cora, Citeseer, and PubMed datasets, respectively. \textit{Red stars} (\textcolor{red}{\ding{72}}) indicate the located GLTs, which reach comparable performance with the least inference MACs. \textit{Dash lines} represent the baseline performance of unpruned GNNs on full graphs. More results over graph sparsity and GNN sparsity are referred to Appendix~\ref{sec:more_small_link}.}
\label{fig:link_pre}
\vspace{-2mm}
\end{figure*}

\begin{table}[htb]
\centering
\vspace{-4mm}
\caption{Graph datasets statistics.}
\label{table:datasets}
\resizebox{0.48\textwidth}{!}{
\begin{tabular}{c |c |c |c |c |c | c | c }
\toprule
Dataset & Task Type &  Nodes & Edges & Ave. Degree & Features & Classes & Metric  \\ 
\midrule
\multirow{2}{*}{Cora} & Node Classification & \multirow{2}{*}{2,708} & \multirow{2}{*}{5,429} & \multirow{2}{*}{3.88} & \multirow{2}{*}{1,433}  & \multirow{2}{*}{7} & Accuracy\\
& Link Prediction & & & & & & ROC-AUC \\ \midrule
\multirow{2}{*}{Citeseer} & Node Classification & \multirow{2}{*}{3,327} & \multirow{2}{*}{4,732} & \multirow{2}{*}{2.84} & \multirow{2}{*}{3,703}  & \multirow{2}{*}{6} & Accuracy \\
& Link Prediction & & & & & & ROC-AUC \\ \midrule
\multirow{2}{*}{PubMed} & Node Classification & \multirow{2}{*}{19,717} & \multirow{2}{*}{44,338} & \multirow{2}{*}{4.50} & \multirow{2}{*}{500}  & \multirow{2}{*}{3} &  Accuracy \\
& Link Prediction & & & & & & ROC-AUC \\ 
\midrule
Ogbn-ArXiv & Node Classification & 169,343 & 1,166,243 & 13.77 & 128 & 40 & Accuracy \\
Ogbn-Proteins & Node Classification & 132,534 & 39,561,252 & 597.00 & 8 & 2 & ROC-AUC\\ \midrule
Ogbl-Collab & Link Prediction & 235,868 & 1,285,465 & 10.90 & 128 & 2 & Hits@50 \\
\bottomrule
\end{tabular}}
\vspace{-2mm}
\end{table}

\vspace{-0.7em}
\section{Experiments}
\vspace{-0.3em}
In this section, extensive experiments are reported to validate the effectiveness of UGS and the existence of GLTs across diverse graphs and GNN models. Our subjects include small- and medium-scale graphs with two-layer Graph Convolutional Network (GCN) \cite{kipf2016semi}, Graph Isomorphism Network (GIN) \cite{xu2018powerful}, and Graph Attention Network (GAT) \cite{velivckovic2017graph} in Section~\ref{sec:large_graph}; as well as large-scale graphs with 28-layer deep ResGCNs \cite{li2020deepergcn} in Section~\ref{sec:large_graph}. Besides, in Section~\ref{sec:pre}, we investigate GLTs under the self-supervised pre-training \cite{you2020graph}. Ablation studies and visualizations are provided in Section~\ref{sec:ablation} and~\ref{sec:vis}.

\vspace{-0.5em}
\paragraph{Datasets} We use popular semi-supervised graph datasets: Cora, Citeseer and PubMed \cite{kipf2016semi}, for both node classification and link prediction tasks. For experiments on large-scale graphs, we use the Open Graph Benchmark (OGB) \cite{hu2020open}, such as Ogbn-ArXiv, Ogbn-Proteins, and Ogbl-Collab. More datasets statistics are summarized in Table~\ref{table:datasets}. Other details such as the datasets' train-val-test splits are included in Appendix~\ref{sec:more_setup}.

\vspace{-0.5em}
\paragraph{Training and Inference Details} Our evaluation metrics are shown in Table~\ref{table:datasets}, following \citet{kipf2016semi,hu2020open,mavromatis2020graph}. More detailed configurations such as learning rate, training iterations, and hyperparameters in UGS, are referred to Appendix~\ref{sec:more_setup}.

\subsection{The Existence of Graph Lottery Ticket} \label{sec:small_graph}
We first examine whether unified graph lottery tickets exist and can be located by UGS. Results of GCN/GIN/GAT on Cora/Citesser/PubMed for node classification and link prediction are collected in Figures~\ref{fig:node_cls} and~\ref{fig:link_pre}, respectively. Note that each point in the figures denotes the achieved performance with respect to a certain graph sparsity, GNN sparsity, and inference MACs. However, due to the limited space, we only include one or two of these three sparsity indicators in the main text, and the rest can be found in Appendix~\ref{sec:more_res}. We list the following \textbf{Obs}ervations.

\vspace{-0.5em}
\paragraph{Obs.1. GLTs broadly exist with substantial MACs saving.} Graph lottery tickets at a range of graph sparsity from $5\%$ to $58.19\%$ without performance deterioration, can be identified across GCN, GIN and GAT on Cora, Citeseer, and PubMed datasets for both node classification and link prediction tasks. Such GLTs significantly reduce $59\%\sim97\%$, $20\%\sim98\%$, $91\%\sim97\%$ inference MACs for GCN, GIN and GAT across all datasets. 

\vspace{-0.5em}
\paragraph{Obs.2. UGS is flexible and consistently shows superior performance.} UGS consistently surpasses random pruning by substantial performance margins across all datasets and GNNs, which validates the effectiveness of our proposal. The previous state-of-the-art method, i.e., ADMM \cite{li2020sgcn}, achieves a competitive performance to UGS at moderate graph sparsity levels, and performs $3\sim4\%$ worse than UGS when graphs are heavily pruned. 

Note that the ADMM approach by \citet{li2020sgcn} is only applicable when two conditions are met: i) graphs are stored via adjacency matrices---however, that is not practical for large graphs \cite{hu2020open}; ii) aggregating features with respect to adjacency matrices---however, recent designs of GNNs (e.g., GIN and GAT) commonly use the much more computation efficient approach of synchronous/asynchronous message passing  \cite{gilmer2017neural,busch2020pushnet}. On the contrary, our proposed UGS is flexible enough and free of these limitations.

\vspace{-0.5em}
\paragraph{Obs.3. GNN-specific and Graph-specific analyses: GAT is more amenable to sparsified graphs; Cora is more sensitive to pruning.} As demonstrated in Figures~\ref{fig:node_cls} and~\ref{fig:link_pre}, compared to GCN and GIN, GLTs in GAT can be found at higher sparsity levels; meanwhile randomly pruned graphs and GAT can still reach satisfied performance and maintain higher accuracies on severely sparsified graphs. 

One possible explanation is that attention-based aggregation is capable of re-identifying important connections in pruned graphs which makes GAT be more amenable to sparsification. Compared the sparsity of located GLTs (i.e., the position of \textit{red stars} (\textcolor{red}{\ding{72}})) across three graph datasets, we find that Cora is the most sensitive graph to pruning and PubMed is more robust to be sparsified. 

\vspace{-2mm}
\subsection{Scale Up Graph Lottery Tickets} \label{sec:large_graph}
To scale up graph lottery tickets, we further conduct experiments on 28-layer deep ResGCNs \cite{li2020deepergcn} on large-scale datasets that have more than millions of connections, like Ogbn-ArXiv and Ogbn-Proteins for node classification, Ogbl-Collab for link prediction in Table~\ref{table:datasets}. We summarize our observations and derive insights below.

\begin{figure}[!ht]
    \centering
    \vspace{-2mm}
    \includegraphics[width=1\linewidth]{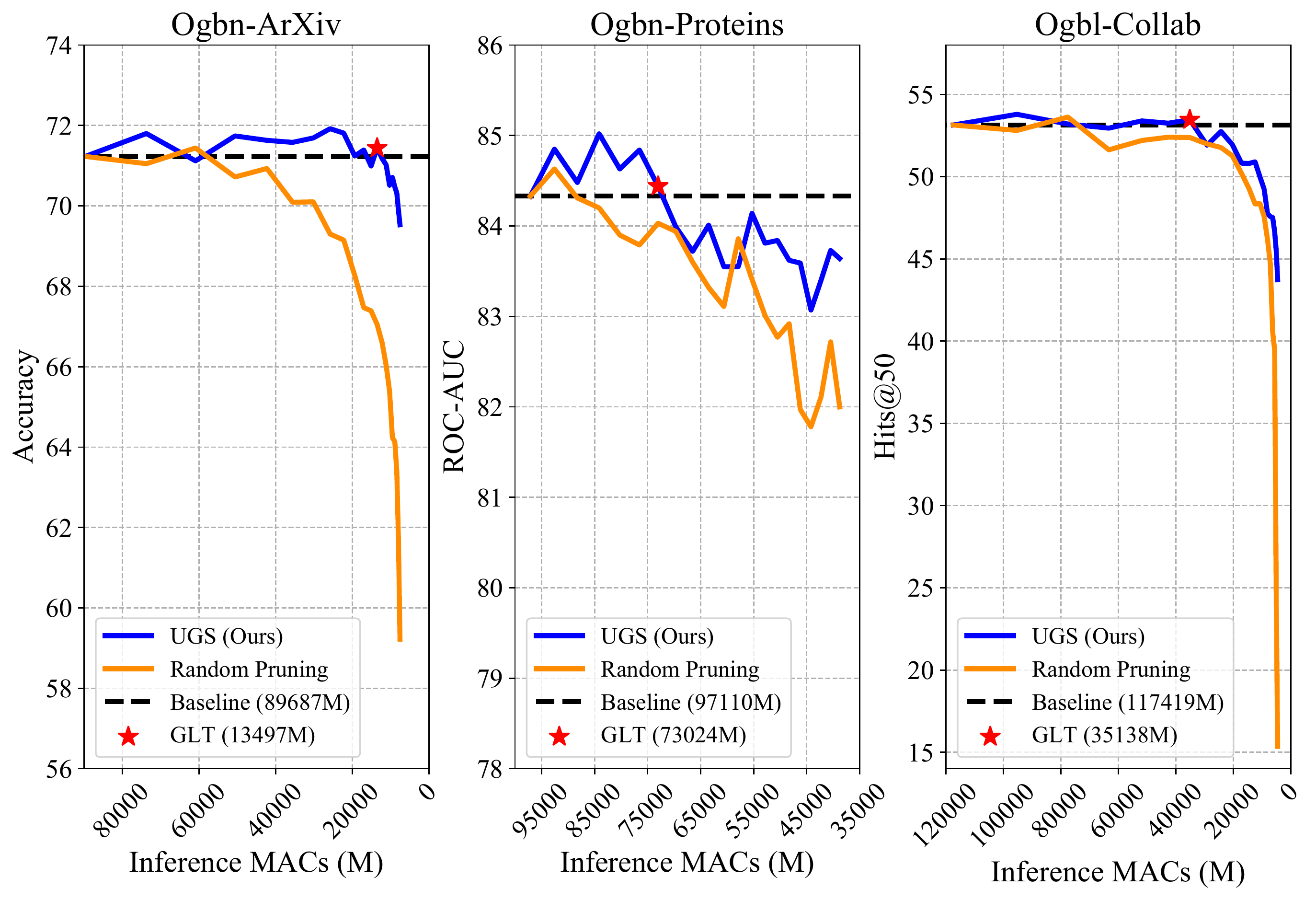}
    \vspace{-8mm}
    \caption{Node classification and link prediction performance of 28-layer \textbf{deep ResGCNs} on \textbf{large-scale graph} datasets.}
    \vspace{-2mm}
    \label{fig:large_gcn}
\end{figure}

\vspace{-0.5em}
\paragraph{Obs.4. UGS is scaleable and GLT exists in deep GCNs on large-scale datasets.} Figure~\ref{fig:large_gcn} demonstrates that UGS can be scaled up to deep GCNs on large-scale graphs. Found GLTs obtain matched performance with $85\%$, $25\%$, $70\%$ MACs saving on Ogbn-ArXiv, Ogbn-Proteins, and Ogbl-Collab, respectively.

\vspace{-0.5em}
\paragraph{Obs.5. Denser graphs (e.g., Ogbn-Proteins) are more resilient to sparsification.} As shown in Figure~\ref{fig:large_gcn}, comparing the node classification results on Ogbn-ArXiv (Ave. degree: 13.77) and Ogbn-Proteins (Ave. degree: 597.00), Ogbn-Proteins has a negligible performance gap between UGS and random pruning, even on heavily pruned graphs. Since nodes with high degrees in denser graphs have less chance to be totally isolated during pruning, it may contribute to more robustness to sparsification. Similar observations can be drawn from the comparison between PubMed and other two small graphs in Figure~\ref{fig:node_cls} and~\ref{fig:link_pre}. 

\vspace{-2mm}
\subsection{Graph Lottery Ticket from Pre-training} \label{sec:pre}

\begin{figure}[!ht]
    \centering
    \vspace{-2mm}
    \includegraphics[width=1\linewidth]{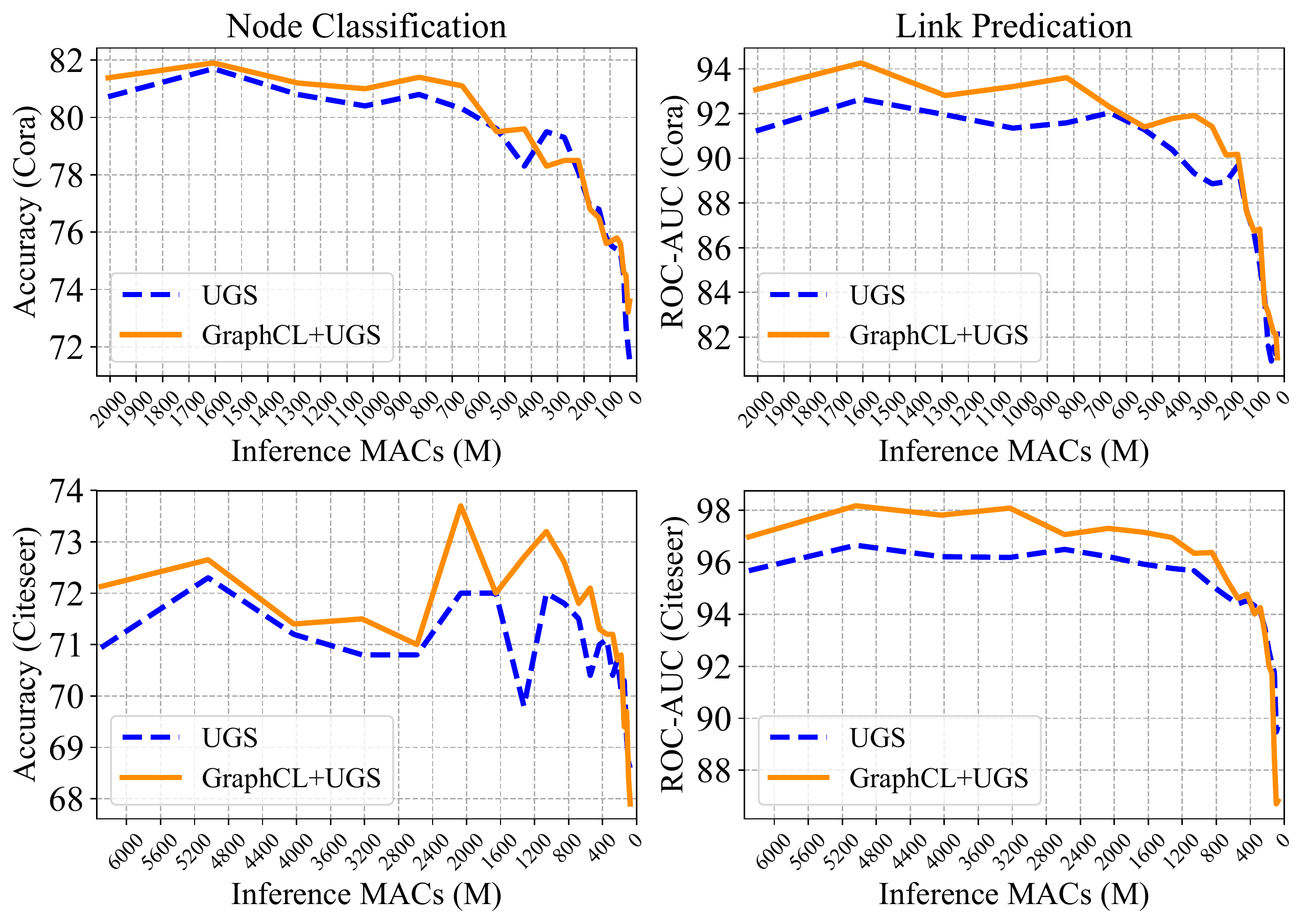}
    \vspace{-8mm}
    \caption{Drawing graph lottery tickets from randomly initialized and self-supervised pre-trained (GraphCL \cite{you2020graph}) GCNs on node classification (low label rate: only $5.0\%$ and $3.6\%$ nodes in Cora and Citesser are labeled) and link prediction.}
    \vspace{-2mm}
    \label{fig:graphcl}
\end{figure}

High-quality lottery tickets can be drawn from self-supervised pre-trained models, as recently found in both NLP and computer vision fields \cite{chen2020lottery,chen2020lottery2}. In the GLT case, we also assess the impact of replacing random initialization with self-supervised graph pre-training, i.e., GraphCL \cite{you2020graph}, on transductive semi-supervised node classification and link prediction. 

From Figure~\ref{fig:graphcl} and~\ref{fig:graphcl_pub}, we gain a few interesting observations. \underline{First}, UGS with the GraphCL pre-trained initialization consistently presents superior performance at moderate sparsity levels ($\le40\%$ graph sparsity $\simeq$ $\le85\%$ MACs saving). While the two settings perform similar at extreme sparsity, it indicates that for excessively pruned graphs, the initialization is no longer the performance bottleneck; \underline{Second}, GraphCL benefits GLT on multiple downstream tasks including node classification and link prediction; \underline{Third}, especially on the transductive semi-supervised setup, GLTs with appropriate sparsity levels can even enlarge the performance gain from pre-training, for example, see GLT on Citeseer with $22.62\%$ graph sparsity and $2068$M inference MACs.

\vspace{-0.3em}
\subsection{Ablation Study} \label{sec:ablation}

\begin{figure}[!ht]
    \centering
    \includegraphics[width=1\linewidth]{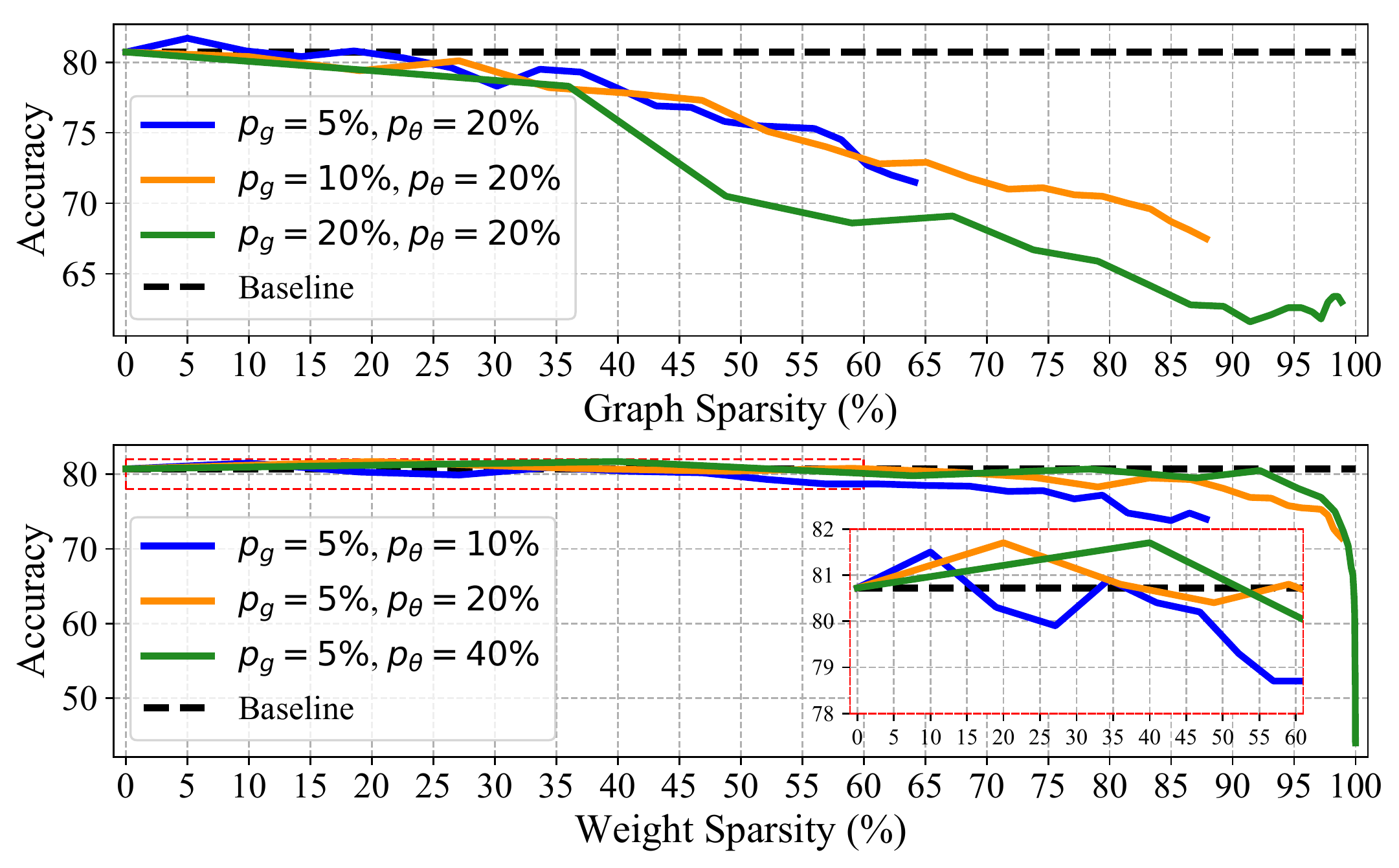}
    \vspace{-8mm}
    \caption{Ablation studies of pruning ratios for the graph and GNN sparsification by UGS, i.e., $p_g,p_{\theta}$. Each curve records the achieved performance of $20$ rounds iterative UGS. GCN on Cora is adopted here. The embedded sub-graph is a zoom-in of the \textcolor{red}{red} box region.}\vspace{-0.5em}
    \label{fig:ablation_ratio}
\end{figure}

\vspace{-0.2em}
\paragraph{Pruning ratio $p_g$ and $p_{\theta}$.} We extensively investigate the pruning ratios $p_g$, $p_{\theta}$ in UGS for graph and GNN sparsification. As shown in Figure~\ref{fig:ablation_ratio}, with a fixed $p_{\theta}=20\%$, only the setting of $p_g=5\%$ can identify the GLT, and it performs close to $p_g=10\%$ at higher sparsity levels (e.g., $\ge25\%$). Aggressively pruning the graph's connections in each round of iterative UGS, e.g., $p_g=20\%$, leads to substantially degraded accuracies, especially for large sparsities. On the other hand, with a fixed $p_{g}=20\%$, all there settings of $p_{\theta}=10\%,20\%,40\%$ show similar performance, and even higher pruning ratios produce slight better results. It again verifies that the key bottleneck in pruning GNNs mainly lies in the sparsification of graphs. In summary, we adopt $p_g=5\%$ and $p_{g}=20\%$ (follow previous LTH works \cite{frankle2018lottery}) for all the experiments.

\begin{table}[!ht]
\centering
  \vspace{-0.5em}
\caption{Performance comparisons of Random GLT versus GLT from GCN on Cora, at several sparsity levels.}
\label{tab:random_GLT}
\resizebox{0.48\textwidth}{!}{
\begin{tabular}{c|cccc}
\toprule
\multirow{2}{*}{Settings} & \multicolumn{4}{c}{(Graph Sparsity, GNN Sparsity)=$(s_g\%,s_{\theta}\%)$} \\ 
\cmidrule{2-5}
& (18.55,59.04) & (22.62,67.23) & (36.98,86.58) & (55.99,97.19) \\ 
\midrule
Random GLT & 79.70 & 78.50 & 75.70 & 63.70 \\
GLT & \textbf{80.80} & \textbf{80.30} & \textbf{79.30} & \textbf{75.30}\\
\bottomrule
\end{tabular}}
\end{table}

\vspace{-0.5em}
\paragraph{Random graph lottery tickets.} Randomly re-initializing located sparse models, i.e., random lottery tickets, usually serves as a necessary baseline for validating the effectiveness of rewinding processes \cite{frankle2018lottery}. In Table~\ref{tab:random_GLT}, we compare GLT to Random GLT, the latter by randomly re-initializing GNN's weights and learnable masks, and GLT shows aapparently superior performance, consistent with previous observations \cite{frankle2018lottery}.

\vspace{-0.3em}
\subsection{Visualization and Analysis} \label{sec:vis}
\begin{figure}[!ht]
    \centering
    \includegraphics[width=0.99\linewidth]{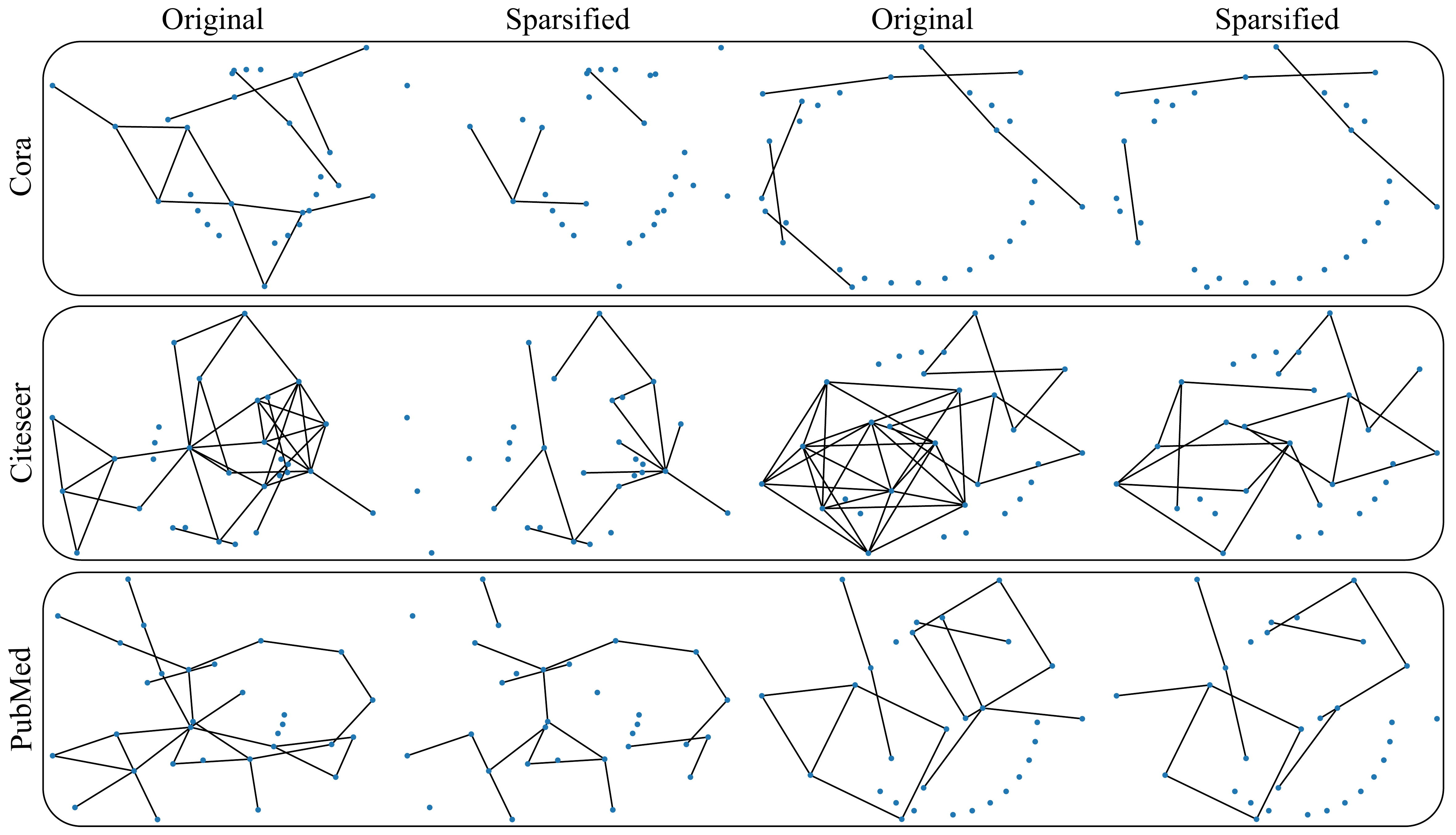}
    \vspace{-4mm}
    \caption{Visualization of sub-graphs (Original/Sparsified) from Cora, Citeseer, and PubMed. Original sub-graphs in the first and third columns are randomly sampled from full graphs. The corresponding unified sparsified sub-graphs of GLTs at $48.67\%$ sparsity, are provided in the second and forth columns.}
    \label{fig:gra_vis}
\end{figure}

\vspace{-0.2em}
In this section, we visualize the sparsified graphs in GLTs from UGS in Figure~\ref{fig:gra_vis}, and further measure the graph properties\footnote{NetworkX ( \href{https://networkx.org}{https://networkx.org}) is used for our analyses.} shown in Table~\ref{tab:compare}, including clustering coefficient~\cite{Luce1949AMO}, as well as node and edge betweenness centrality~\cite{freeman1977set}. Specifically, clustering coefficient measures the proportion of edges between the nodes within a given node’s neighborhood; node and edge betweenness centrality show the degree of $\textit{central}$ a vertex or an edge is in the graph~\cite{narayanan2005betweenness}. Reported numbers in Table~\ref{tab:compare} are averaged over all the nodes.

Both Figure~\ref{fig:gra_vis} and Table~\ref{tab:compare} show that sparse graphs obtained from UGS seem to maintain more ``critical" vertices which used to have much denser connections. It may provide possible insights on what GLTs prefer to preserve.

\vspace{-0.1em}
\section{Conclusion and Discussion}
\vspace{-0.18em}
In this paper, we first propose unified GNN sparsification to generalize the notion or pruning in GNNs. We further establish the LTH for GNNs, by leveraging UGS and considering a novel joint data-model lottery ticekt. The new unified LTH for GNNs generalizes across various GNN architectures, learning tasks, datasets, and even initialization ways. In general, we find GLT to tremendously trim down the inference MACs, without sacrificing task performance. 


It remains open how much we could translate GLT's high sparsity into practical acceleration and energy-saving benefits. Most DNN accelerators  are optimized for dense and regular computation, making edge-based operations hard to implement efficiently. To our best knowledge, the hardware acceleration research on GNNs just starts to gain interests \cite{auten2020hardware,abadal2020computing,geng2020awb,wang2020gnn,kiningham2020grip}. We expect GLT to be implemented using sparse-dense matrix multiplication (SpMM) operations from highly optimized sparse matrix libraries, such as Intel MKL \cite{wang2014intel} or cuSPARSE \cite{naumov2010cusparse}. 


\clearpage

\bibliography{GraphUS}
\bibliographystyle{icml2021}

\clearpage

\appendix
\renewcommand{\thepage}{A\arabic{page}}  
\renewcommand{\thesection}{A\arabic{section}}   
\renewcommand{\thetable}{A\arabic{table}}   
\renewcommand{\thefigure}{A\arabic{figure}}

\section{More Implementation Details} \label{sec:more_setup}
\paragraph{Datasets Download Links.}
As for small-scale datasets, we take the commonly used semi-supervised node classification graphs: Cora, Citeseer and pubMed. For larger-scale datasets, we use three Open Graph Benchmark (OGB) \cite{hu2020open} datasets: Ogbn-ArXiv, Ogbn-Proteins and Ogbl-Collab. All the download links of adopted graph datasets are included in Table~\ref{table:datasets_link}.   

\begin{table}[htb]
\centering
\caption{Download links of graph datasets.}
\label{table:datasets_link}
\resizebox{0.48\textwidth}{!}{
\begin{tabular}{c |c }
\toprule
Dataset & Download links and introduction websites  \\ 
\midrule
Cora & https://linqs-data.soe.ucsc.edu/public/lbc/cora.tgz \\
Citeseer & https://linqs-data.soe.ucsc.edu/public/lbc/citeseer.tgz  \\
PubMed & https://linqs-data.soe.ucsc.edu/public/Pubmed-Diabetes.tgz \\
\midrule
Ogbn-ArXiv & https://ogb.stanford.edu/docs/nodeprop/\#ogbn-arxiv \\
Ogbn-Proteins & https://ogb.stanford.edu/docs/nodeprop/\#ogbn-proteins \\
Ogbl-Collab & https://ogb.stanford.edu/docs/linkprop/\#ogbl-collab \\
\bottomrule
\end{tabular}}
\vspace{-4mm}
\end{table}

\paragraph{Train-val-test Splitting of Datasets.} As for node classification of small- and medium-scale datasets, we use 140 (Cora), 120 (Citeseer) and 60 (PubMed) labeled data for training, 500 nodes for validation and 1000 nodes for testing.  As for link prediction task of small- and medium-scale datasets Cora, Citeseer and PubMed, we random sample $10\%$ edges as our testing set, $5\%$ for validation, and the rest $85\%$ edges are training set. The training/validation/test splits for Ogbn-ArXiv, Ogbn-Proteins and Ogbl-Collab are given by the benchmark \cite{hu2020open}. Specifically, as for Ogbn-ArXiv, we train on the papers published until 2017, validation on those published in 2018 and test on those published since 2019. As for Ogbn-Proteins, we split the proteins nodes into training/validation/test sets according to the species which the proteins come from. 
As for Ogbl-Collab, we use the collaborations until 2017 as training edges, those in 2018 as validation edges, and those in 2019 as test edges.

\paragraph{More Details about GNNs.}

As for small- and medium-scale datasets Cora, Citeseer and PubMed, we choose the two-layer GCN/GIN/GAT networks with 512 hidden units to conduct all our experiments.
As for large-scale datasets Ogbn-ArXiv, Ogbn-Proteins and Ogbl-Collab, we use the ResGCN \cite{li2020deepergcn} with 28 GCN layers to conduct all our experiments. As for Ogbn-Proteins dataset, we also use the edge encoder module, which is a linear transform function in each GCN layer to encode the edge features. And we found if we also prune the weight of this module together with other weight, it will seriously hurt the performance, so we do not prune them in all of our experiments.

\paragraph{Training Details and Hyper-parameter Configuration.} 
We conduct numerous experiments with different hyper-parameter, such as iterations, learning rate, $\gamma_1$, $\gamma_2$, and we choose the best hyper-parameter configuration to report the final results.
All the training details and hyper-parameters are summarzed in Table \ref{table:train_detials}.
As for Ogbn-Proteins dataset, due to its too large scale, we use the commonly used random sample method \cite{li2020deepergcn} to train the whole graph. 
Specifically, we random sample ten subgraphs from the whole graph and we only feed one subgraph to the GCN at each iteration. For each subgraph, we train 10 iterations to ensure better convergence. And we train 100 epochs for the whole graph (100 iterations for each subgraph).

\paragraph{Evaluation Details}
We report the test accuracy/ROC-AUC/Hits@50 according to the best validation results during the training process to avoid overfitting. All training and evaluation are conducted for one run. As for the previous state-of-the-art method ADMM \cite{li2020sgcn}, we use the same training and evaluation setting as the original paper description, and we can reproduce the similar results compared with original paper.    

\paragraph{Computing Infrastructures}

We use the NVIDIA Tesla V100 (32GB GPU) to conduct all our experiments.

\begin{table*}[htb]
\centering
\caption{Implementation details of node classification and link prediction.}
\label{table:train_detials}
\resizebox{0.98\textwidth}{!}{
\begin{tabular}{c |c |c |c | c | c | c | c | c | c}
\toprule
Task & \multicolumn{5}{c|}{Node Classification} & \multicolumn{4}{c}{Link Prediction} \\ \cmidrule{2-6} \cmidrule{7-10}
Dataset & Cora & Citeseer & PubMed & Ogbn-ArXiv & Ogbn-Proteins & Cora & Citeseer & PubMed & Ogbn-Collab\\
\midrule
Iteration & 200 & 200 & 200 & 500 & 100 &200 & 200 & 200 & 500\\
Learning Rate & 8e-3 & 1e-2 & 1e-2 & 1e-2 & 1e-2& 1e-3 & 1e-3& 1e-3 & 1e-2\\
Optimizer & Admm & Admm & Admm & Admm & Admm & Admm & Admm & Admm & Admm\\
Weight Decay & 8e-5 & 5e-4 & 5e-4 & 0 & 0 & 0 & 0 & 0 & 0 \\
$\gamma_1$ & 1e-2 & 1e-2 & 1e-6 & 1e-6 & 1e-1 & 1e-4 & 1e-4 & 1e-4 & 1e-6  \\
$\gamma_2$  & 1e-2 & 1e-2 & 1e-3 & 1e-6 & 1e-3 & 1e-4 & 1e-4 & 1e-4 & 1e-5 \\
\bottomrule
\end{tabular}}
\end{table*}

\section{More Experiment Results} \label{sec:more_res}

\subsection{Node Classification on Small- and Medium-scale Graphs with shallow GNNs} \label{sec:more_small_node}

\begin{figure*}[!ht]
    \centering
    \includegraphics[width=1\linewidth]{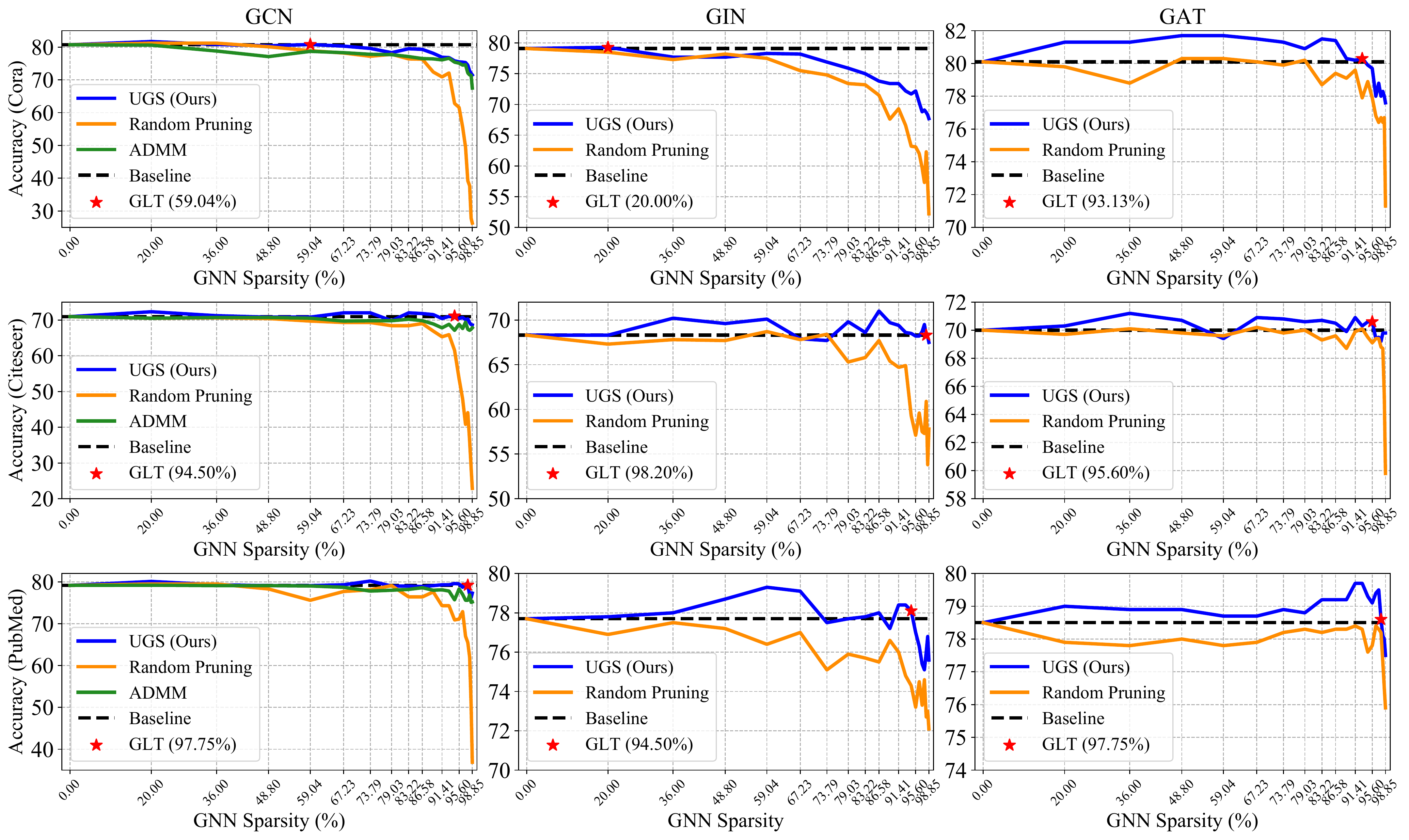}
    \vspace{-8mm}
    \caption{\textbf{Node classification} performance over achieved GNNs sparsity of GCN, GIN, and GAT on Cora, Citeseer, and PubMed datasets, respectively. \textit{Red stars} (\textcolor{red}{\ding{72}}) indicate the located GLTs, which reach comparable performance with extreme GNN sparsity. \textit{Dash lines} represent the baseline performance of unpruned GNNs on full graphs.}
    \label{fig:more_small_weight}
\end{figure*}

As shown in Figure~\ref{fig:more_small_weight}, we also provide extensive results over GNN sparsity of GCN/GIN/GAT on three small datasets, Cora/Citeseer/PubMed. We observe that UGS finds the graph wining tickets at a range of GNNs sparsity from $20\%\sim90\%$ without performance deterioration, which significantly reduces MACs and the storage memory during both training and inference processes. 

\subsection{Link Prediction on Small- and Medium-scale Graphs with shallow GNNs} \label{sec:more_small_link}

\begin{figure*}[!ht]
    \centering
    \includegraphics[width=1\linewidth]{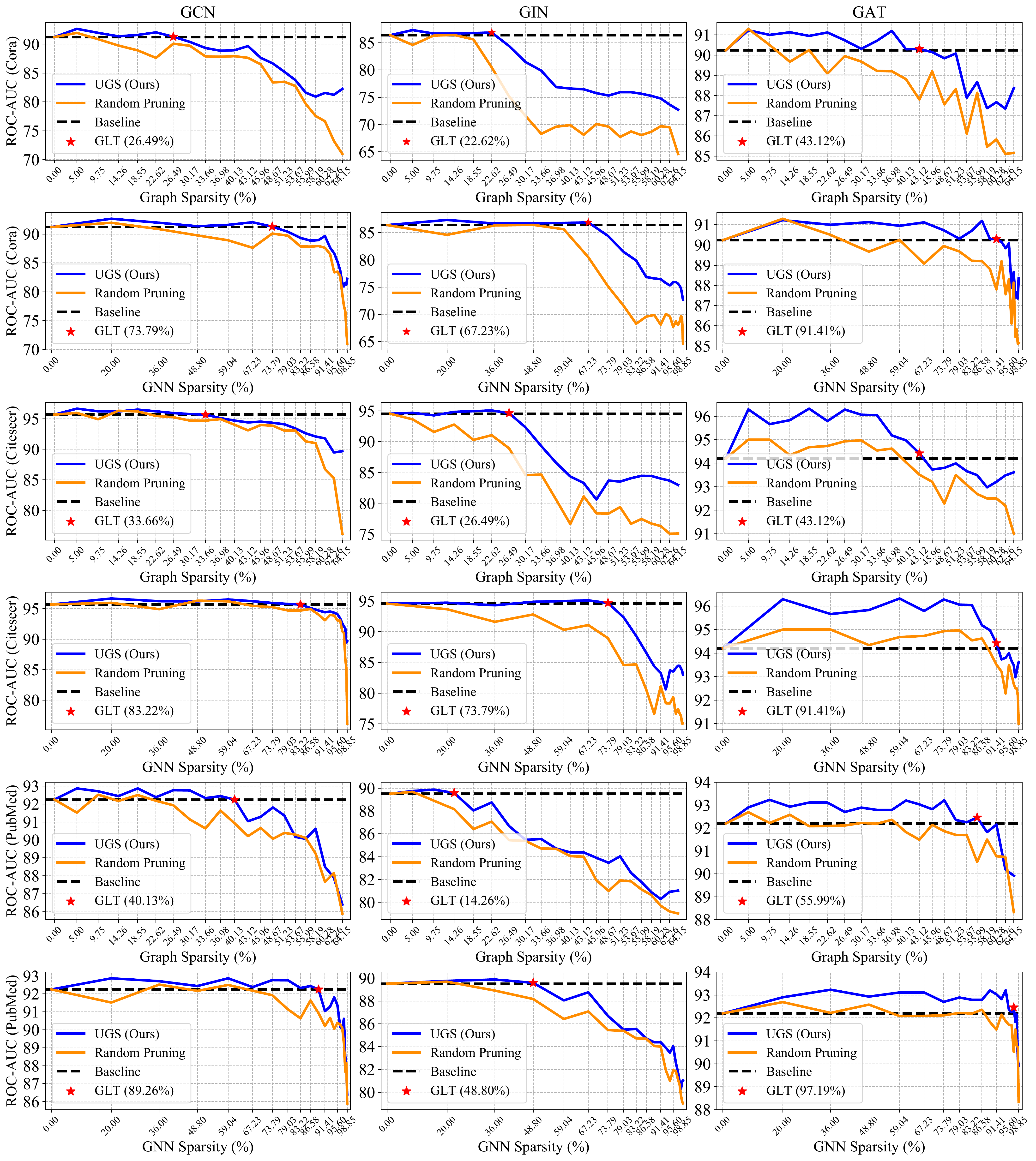}
    \vspace{-8mm}
    \caption{\textbf{Link prediction} performance over achieved GNNs sparsity of GCN, GIN, and GAT on Cora, Citeseer, and PubMed datasets, respectively. \textit{Red stars} (\textcolor{red}{\ding{72}}) indicate the located GLTs, which reach comparable performance with the extreme graph sparsity and GNN sparsity. \textit{Dash lines} represent the baseline performance of unpruned GNNs on full graphs.}
    \label{fig:more_small_link}
\end{figure*}

More results of link prediction with GCN/GIN/GAT on Cora/Citeseer/PubMed datasets are shown in Figure~\ref{fig:more_small_link}. We observe the similar phenomenon as the node classification task: using our proposed UGS can find the graph wining tickets at a range of graph sparsity from $5\%\sim50\%$ and GNN sparsity from $20\%\sim90\%$ without performance deterioration, which greatly reduce the computational cost and storage space during both training and inference processes.

\subsection{Large-scale Graphs with Deep ResGCNs} \label{sec:more_large}
\begin{figure*}[!ht]
    \centering
    \includegraphics[width=1\linewidth]{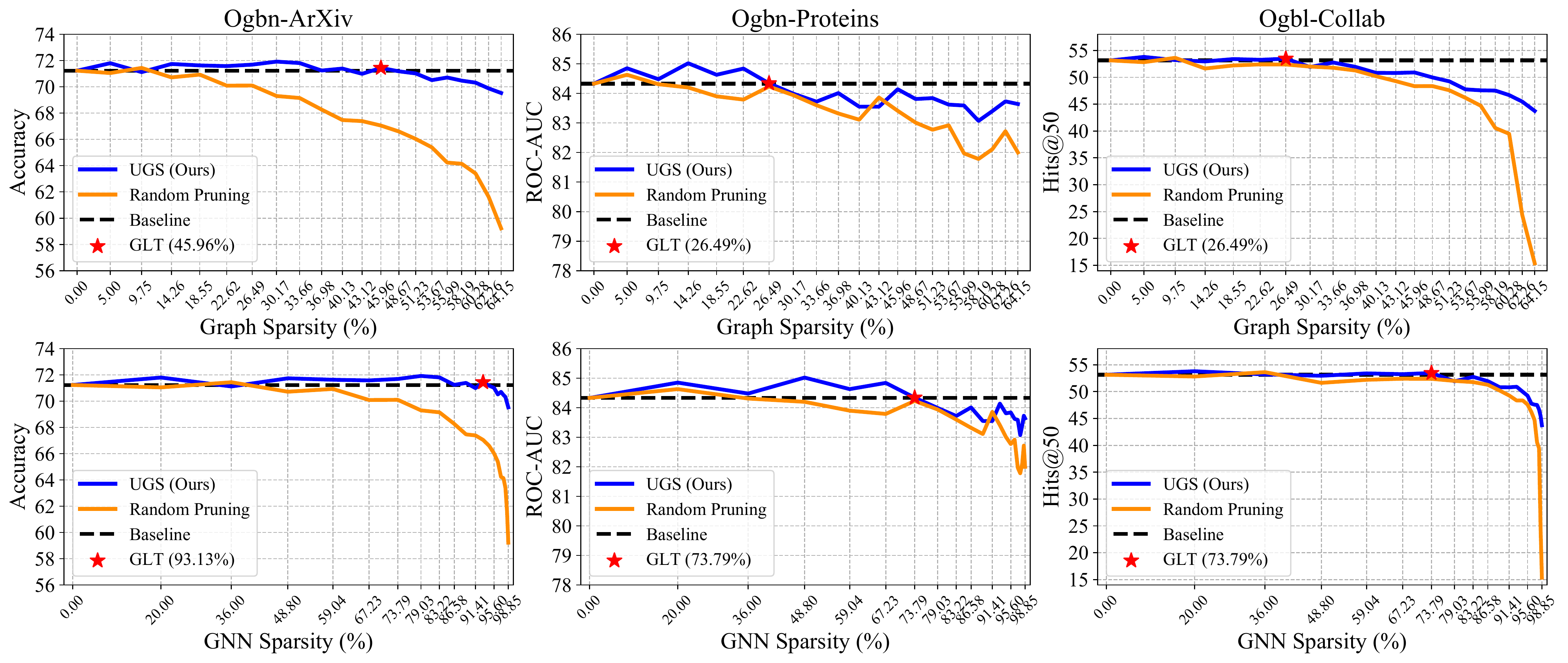}
    \vspace{-8mm}
    \caption{Node classification and link prediction performance over achieved graph sparsity and GNN sparsity of 28-layer \textbf{deep ResGCNs} on \textbf{large-scale graph} datasets.}
    \label{fig:more_large_gcn}
\end{figure*}

More results of larger-scale graphs with deep ResGCNs are shown in Figure~\ref{fig:more_large_gcn}. Results show that our proposed UGS can found GLTs, which can reach the non-trivial sparsity levels of graph $30\% \sim 50\%$ and weight $20\% \sim 80\%$ without performance deterioration. 

\subsection{Graph Lottery Ticket with Pre-training} 

\begin{figure*}[!ht]
    \centering
    \includegraphics[width=1\linewidth]{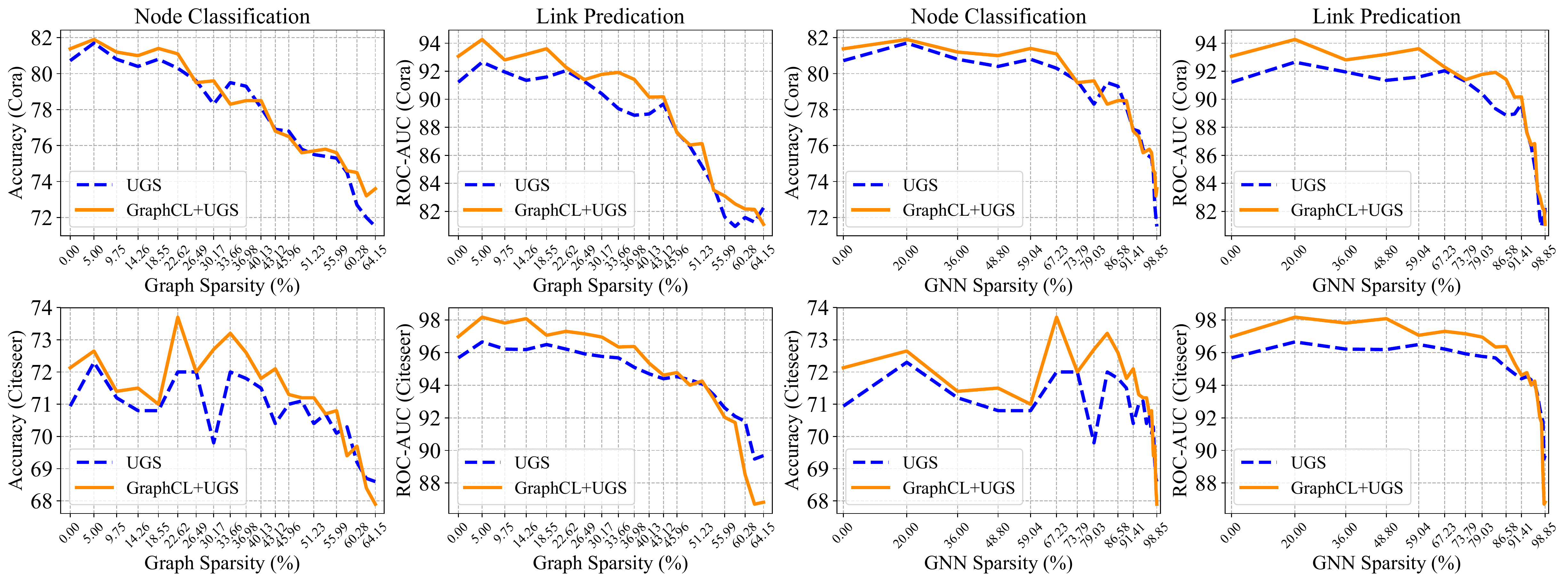}
    \vspace{-8mm}
    \caption{Drawing graph lottery tickets from randomly initialized and self-supervised pre-trained (GraphCL \cite{you2020graph}) GCNs on node classification (low label rate) and link prediction. Corresponding results over achieved graph and GNN sparsity are presented.}
    \label{fig:graphcl_pub}
\end{figure*}

More results of node classification and link prediction on Cora and Citeseer dataset of GraphCL \cite{you2020graph} pre-training are shown in Figure \ref{fig:graphcl_pub}. Results demonstrate that when using self-supervised pre-training, UGS can identify graph lottery tickets with higher qualities. 

\subsection{More Analyses of Sparsified Graphs}
As shown in Table~\ref{tab:compare}, graph measurements are reported, including clustering coefficient, node and egde betweeness centrality. Results indicate that UGS seems to produce sparse graphs with more ``critical" vertices which used to have more connections.

\begin{table*}[!ht]
\centering
\caption{Graph measurements of original graphs, sparse graphs from UGS, random pruning, and ADMM sparsification.}
\label{tab:compare}
\resizebox{0.98\textwidth}{!}{
\begin{tabular}{c|c|ccc|c|ccc|c|ccc}
\toprule
\multirow{2}{*}{Measurements} & \multicolumn{4}{c|}{Cora} & \multicolumn{4}{c|}{Citeseer} & \multicolumn{4}{c}{PubMed} \\ 
\cmidrule{2-13}
& Original & UGS & RP & ADMM & Original & UGS & RP & ADMM & Original & UGS & RP & ADMM \\ 
\midrule
Clustering Coefficient & 0.14147 & 0.07611 & \textbf{0.08929} & 0.04550 &  0.24067 & \textbf{0.15855} & 0.15407 & 0.08609 & 0.06018 & 0.03658 & \textbf{0.03749} & 0.02470 \\
Node Betweenness & 0.00102 & \textbf{0.00086} & 0.00066 & 0.00021 & 0.00165 & \textbf{0.00190} & 0.00158 & 0.00122 & 0.00027 & \textbf{0.00024} & 0.00022 & 0.00018 \\
Edge Betweenness & 0.00081 & \textbf{0.00087} & 0.00068 & 0.00032 & 0.00101 & \textbf{0.00144} & 0.00123 & 0.00137 & 0.00014 & \textbf{0.00019} & 0.00015 & 0.00018  \\
\bottomrule
\end{tabular}}
\end{table*}

\end{document}